\documentclass[letterpaper,journal]{IEEEtran}
\usepackage{amsmath,amsfonts}
\usepackage{amssymb}
\usepackage{algorithmic}
\usepackage{algorithm}
\usepackage{array}
\usepackage[caption=false,font=footnotesize]{subfig}
\usepackage{textcomp}
\usepackage{stfloats}
\usepackage{url}
\usepackage{graphicx}
\usepackage{cite}
\usepackage{tabularx}
\usepackage{graphicx,tikz}
\usepackage{graphicx}
\usepackage{adjustbox} 
\usepackage{booktabs}
\usetikzlibrary{positioning, arrows.meta, calc}

\begin{document}

\title{Acoustic Signature Optimisation for Bioinspired Flapping-Wing Micro Aerial Vehicles}

\author{Yi Lun Duh, Krispin C. V. Broers and Sophie F. Armanini}

\markboth{Preprint October 2026}{}

\maketitle

\begin{abstract}
Flapping-wing micro air vehicles are prized for quiet flight, yet the few existing measurements show engineered wings to be 15--30\,dB louder than the animals they imitate.
This paper investigates two passive geometric approaches to reduce the acoustic signature of a flapping membrane wing, and applies these to an exemplary wing.
A significant sound source in flapping wings is the impulsive click radiated at stroke reversal, when a slack membrane snaps between two curvatures. Our first noise-reduction approach optimises the wing's venation layout, based on an analytical model of the reversal click as a dynamic snap-through of the membrane. 
The optimised venation pattern was manufactured and resulted in a reduction of up to 8\,dB during the reversal snap in experiments. 
During testing, the venation was found to be sensitive to orientation rather than thickness.
Secondly, an additive trailing-edge fringe inspired by hummingbird feathers was shown to further reduce the broadband swoosh by 3--4\,dB at no lift cost, matching a cut serration wing without removing lifting area.
\end{abstract}

\section{Introduction}\label{sec:introduction}
\IEEEPARstart{M}{icro} aerial vehicles are increasingly deployed in confined, cluttered spaces shared with people and wildlife, where safety and unobtrusiveness matter as much as endurance. Flapping-wing micro aerial vehicles (FWMAVs) suit such settings well, because their wings are light and compliant and the wingtips slow to a stop at each stroke reversal, so a collision carries far less energy than the fast, stiff rotor of a comparable multirotor~\cite{rafee_nekoo_review_2025}. This ability to operate safely in tight spaces makes FWMAVs attractive for tasks such as infrastructure inspection, wildlife observation and indoor monitoring. Tasks that operate close to animals or people, in particular, benefit directly from quiet flight and an unobtrusive vehicle. Low acoustic emission is repeatedly cited as a key advantage of flapping over rotary-wing flight, yet the claim is rarely tested. Across the FWMAV literature, only a handful of studies measure noise at all~\cite{keennon_development_2012,lu_bioinspired_2018,debiasi_low-noise_2020,wu_experimental_2026}, and two recent comprehensive FWMAV reviews compare flight time, mass, mechatronics, control and autonomy, but neither reports the acoustic signature of a single platform~\cite{rafee_nekoo_review_2025,hashim_insects_2026}.

\begin{figure}[t]
    \centering
    \includegraphics[width=0.9\linewidth]{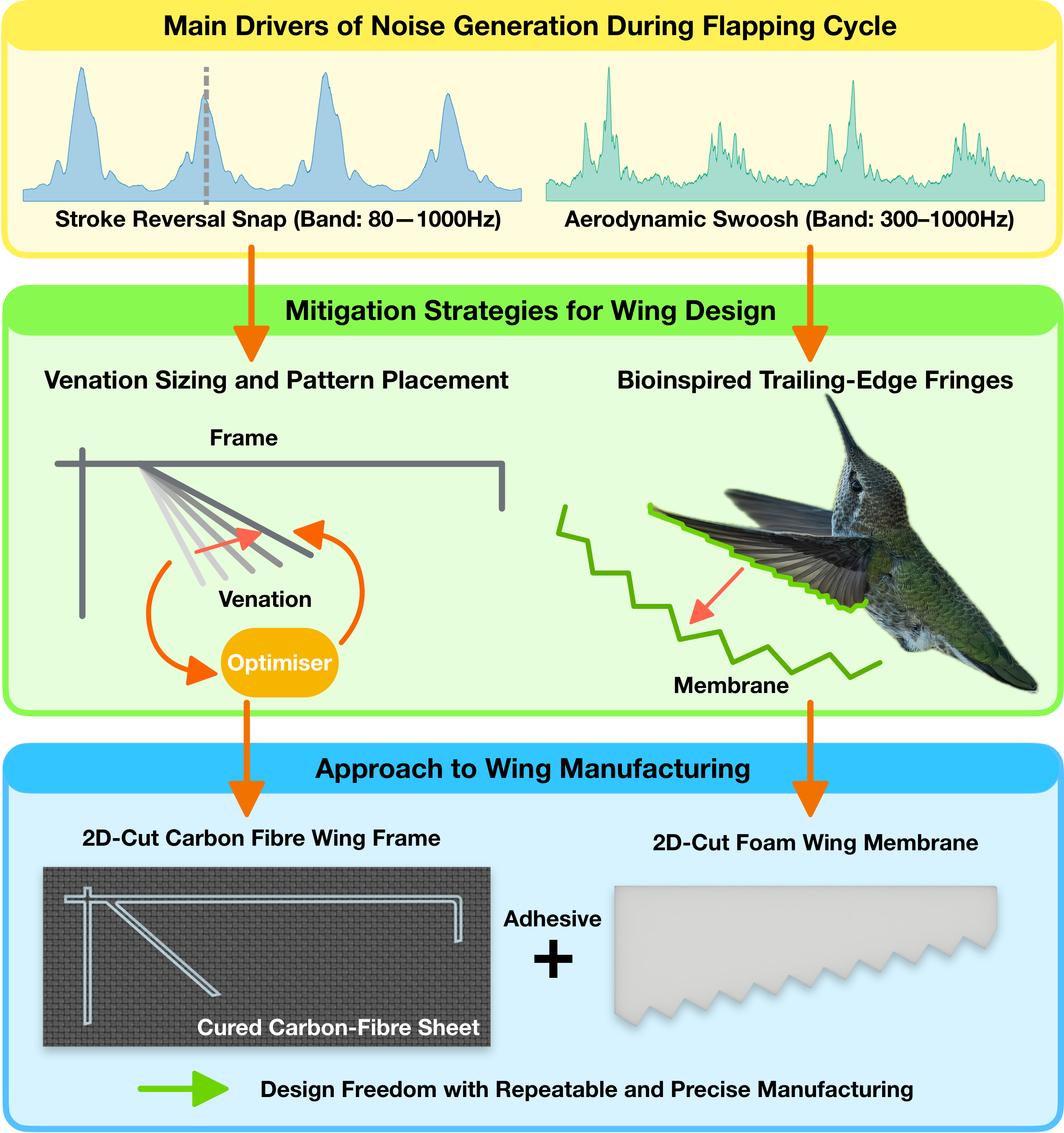}
    \caption{Wing design framework for acoustically optimising flapping-wing robots.}
    \label{fig:framework}
\end{figure}

The few available measurements contradict the assumption of quietness~\cite{rafee_nekoo_review_2025,keennon_development_2012,lu_bioinspired_2018}. E.g., a Mylar-membrane wing on carbon rods was found to emit 68.8\,dBA at 10\,Hz in Ref.~\cite{lu_bioinspired_2018}, and the Nano Hummingbird produced 79.7\,dBA at 42.5\,Hz, compared to the 40--50\,dB produced by a live hummingbird~\cite{keennon_development_2012}. Engineered flapping wings are therefore 15--30\,dB louder than the animals they imitate. Keennon et al.\ also found, that most of the noise is produced at stroke reversal, when the large angular acceleration makes the membrane ``pop through from the fore-stroke configuration to the aft-stroke configuration in an unstructured and violent manner,'' likened to a sail snapping as a yacht tacks~\cite{keennon_development_2012}. This impulsive event, which we call the reversal snap, occurs twice per cycle and was reaffirmed by Wu et al.~\cite{wu_experimental_2026}, yet no subsequent design has set out to suppress it.

How a wing turns its motion into sound explains why this event is so loud and how it might be quietened. During flapping, an unsteady surface force pushes on the surrounding air. The time derivative of that force radiates to the far field~\cite{curle_1955}. A wing is quiet when its push on the air is smooth and loud when that push changes abruptly, as it does when the membrane snaps through. Clark~\cite{clark_ways_2021} catalogues several mechanisms by which a flapping wing emits sound, and two of these dominate for a membrane FWMAV.

The first is the broadband swoosh radiated by the vortical flow shed near the trailing edge, which is controlled by the shape of that edge, or its serration.
The second is the bistable snap, whose radiated sound is the atonal reversal click, produced when a slack membrane cannot carry the reversal inertia in tension and instead inverts its curvature. The strength of the click is set by the bending stiffness of the membrane and by how that stiffness is distributed, which a layout of bonded carbon stiffeners, known as the wing venation, can reshape. 

Existing work suggests that the membrane and the edge both influence the noise, but existing work has not attempted to use either the venation layout or the edge geometry to reduce noise. 
Lu et al.~\cite{lu_bioinspired_2018} cut perceived noise by 12\,dBA by replacing the Mylar membrane with a dielectric elastomer, and Debiasi et al.~\cite{debiasi_low-noise_2020} reduced membrane rustling by tensioning it with bonded stiffeners, a structural effect since a tensioned sheet cannot snap through. Wu et al.~\cite{wu_experimental_2026} provide the most detailed measurement to date, finding the wings rather than the transmission to dominate and to radiate as a stroke-plane dipole. They assign low-, mid- and high-frequency content to aerodynamic loading, transmission resonance and membrane flutter with vortex shedding, with the pitching phase near reversal identified as the primary noise generator.

These two design variables are themselves well studied, but for performance rather than for sound. Insect wings are stiffened membranes whose venation sets how a wing of a given size distributes its stiffness~\cite{combes_flexural_2003}, and flapping freshly cut insect wings in helium show that fluid forces barely affect wing bending, which supports a damped structural model with no fluid loading~\cite{combes_into_2003}. Building on this, venation and graded rigidity have been optimised for lift, thrust and efficiency~\cite{stanford_aeroelastic_2012,reade_investigation_2022,zhu_parametric_2025,truong_effects_2026}, but never as an acoustic variable. Edge serrations are equally well characterised on rigid geometries, where the fixed-wing airfoil literature shows sawtooth cuts reducing broadband scattering noise by a few decibels, improving with tooth height until they saturate~\cite{gruber_experimental_2013,avallone_benefits_2017}, and Ji et al.~\cite{ji_influences_2022} report up to 6.8\,dB reduction on a rigid flapping-wing in hover. Serrations on a flexible membrane wing under flapping loads have, to our knowledge, not been studied. The reversal click and the trailing-edge swoosh are therefore identified noise sources, and the venation layout and the edge serration are geometric variables that could be set to reduce the noise sources without active actuation.

In this paper, we propose a wing design framework (see~Fig.~\ref{fig:framework}) for acoustically optimising flapping wings utilising an integrated wing layout manufacturing process.
This process enables the implementation of a venation layout optimisation loop, which is combined with an experimentally optimised trailing-edge fringe design, addressing both main sources of noise in the flapping wing.
Our main contributions to lowering the acoustic signature of flapping-wing robots are the following:

\begin{itemize}
\item Venation layout optimisation model for improved acoustic signature during stroke reversal.
\item Bioinspired trailing-edge fringes as an additive membrane pattern reducing the wing's aerodynamic swoosh noise.
\item Integrated manufacturing approach allowing design freedom in wing shape and venation layout.
\item Experimental investigation of the proposed wing design framework using a microphone and load cell test setup.
\end{itemize}

The remainder of the paper is organised as follows: In Sec.~\ref{sec:framework}, the proposed wing design framework and its strategies for noise reduction are described. Sec.~\ref{sec:venation-model} explains the architecture of the acoustic optimisation model. The experimental evaluation of the mitigation strategies is then presented in Sec.~\ref{sec:venation-results}, followed by the conclusions in Sec.~\ref{sec:conclusion}.

\section{Wing Design Framework}\label{sec:framework}
This chapter details the proposed wing design framework and its strategies to reduce the sounds generated by the flapping wings.

\subsection{Design and Manufacturing}
Most FWMAVs feature a sail-like wing architecture using carbon fibre rods for the leading edge and root chord spanning a thin and flexible membrane made from foam or foil sheets. This membrane is often reinforced by venation ribs acting as stiffeners, which are commonly glued onto the membrane (e.g.~\cite{keennon_development_2012,preumont_note_2021}). This construction limits the available wing design space and requires connecting structures to rigidly bond the individual carbon fibre rods together, increasing complexity and weight. Furthermore, when experimentally testing wing design variations, manufacturing multiple designs with high precision can be challenging. 
We therefore implemented an integrated manufacturing process supporting the proposed venation layout optimisation and allowing the repeatable production of precise wing designs.

This manufacturing process combines two components: a cured carbon fibre sheet based wing frame with an integrated venation layout which is bonded via an adhesive layer to a foam-sheet based membrane cut-out.
Instead of connecting several precisely cut carbon fibre rods to achieve a desired wing layout, we utilised a CNC waterjet cutting machine to produce a single-piece wing frame with integrated venation patterns from a cured carbon fibre sheet.
This allows the realisation of novel wing designs and the efficient optimisation of features, such as the venation pattern explored in this paper.

To demonstrate the utility of our approach, a baseline wing is produced for application and evaluation in a previously developed \cite{broers_multimodal_2026} flapping-mechanism test rig (see Fig.~\ref{fig:jig}):
\begin{figure}[htb]
    \centering
    \includegraphics[width=0.7\columnwidth]{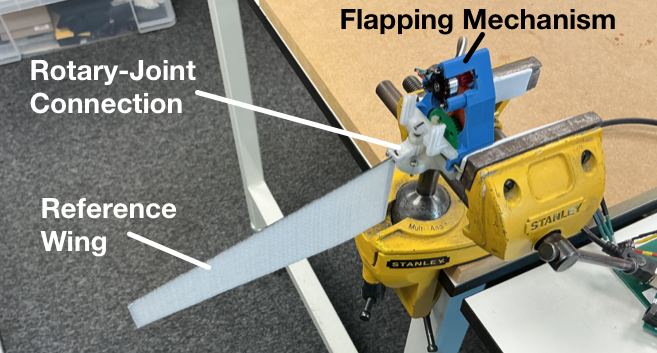}
    \caption{Passive-pitch flapping mechanism \cite{broers_multimodal_2026} featuring the baseline wing design B0.}
    \label{fig:jig}
\end{figure}
This wing provides a mating feature at the intersection of its root chord and leading edge, connecting it to a rotary joint on the flapping mechanism, allowing passive wing pitch around the leading-edge axis. The mating feature is friction-fit with a thin-walled steel tube, which rotates against a sleeve joint that is oscillated in a wide angle by the flapping mechanism. Additionally, a small extension piece of the root chord crossing the leading edge acts as a mechanical limiter to this rotary joint by hitting two defined end-stops. This passive wing-rotation system precisely defines the wing's maximum pitch angle during its translation phase of the flapping cycle, while facilitating a frictionless wing reversal during the rotation phase.
Another unique feature of this integrated wing design is the inclusion of a tip chord allowing for a larger supported wing area with a shorter wing span compared to a triangular wing (e.g.~\cite{wu_design_2026}). As this stiff tip chord is rigidly connected to the leading edge, it produces a passive wing-twist profile during the translation phase, dependent on the torsional stiffness of the leading edge.
This baseline wing frame features a simple membrane design with a straight trailing edge and no venation pattern. For the planform a tapered trapezium of span $L = 135\,\mathrm{mm}$, root chord $c_r = 45\,\mathrm{mm}$, and tip chord $c_t = 15\,\mathrm{mm}$ (taper ratio 0.33) was chosen. Due to the manufacturing process from a single cured carbon fibre sheet ($E_{CF} = 70\,\mathrm{GPa}$, $\rho_{CF} = 1300\,\mathrm{kg/m^3}$), the wing frame features a unified thickness of 1\,mm. The exact 2D frame geometry, for example, a root chord width profile, can be freely designed,  defining the stiffness properties of the wing.
For the membrane material, expanded-polyethylene (EPE) wrap with an LDPE base resin was chosen with a total thickness of 0.7\,mm ($E_m = 5.0\,\mathrm{MPa}$, $\rho_{M} = 25\,\mathrm{kg/m^3}$). This membrane was bonded on one face of the wing frame using a UHU~POR adhesive. Lastly, following a curing time the desired planform was cut out with a scalpel.
This manufacturing process was repeated for all wings tested in this paper.

\subsection{Bioinspired Trailing-Edge Fringes}
The first mitigation strategy investigated in this paper is the utilisation of trailing-edge fringes inspired by the feather pattern found in a hummingbird with extended wings.
These fringes address the broadband sound source of the aerodynamic swoosh radiated near the trailing edge when shed vorticity imposes fluctuating pressure on the wing. Integrated over the wing, this is an unsteady force whose time derivative radiates by the same Curle relation as the snap~\cite{ji_influences_2022}, but its content sits higher in frequency and is set by the flow. Wu et al.~\cite{wu_experimental_2026} place it in 300--1000\,Hz, so the fringe is scored in that swoosh band, the same band over which Ji et al.~\cite{ji_influences_2022} deliver their reduction.

The fixed-wing literature favours the pointed (triangular) tooth over the blunt rectangular one, because a blunt root gap re-radiates a coherent vortex pair~\cite{gruber_experimental_2013,avallone_benefits_2017}, so the triangular tooth is adopted here. A serration \emph{cut} into the membrane removes lifting area at the free edge and opens a tear path at the tooth roots. An \emph{additive} fringe avoids both, keeping the baseline planform intact and adding triangular teeth beyond the mean trailing edge, so no load-bearing membrane is removed. The precedent is Preumont's robotic hummingbird~\cite{preumont_note_2021}, whose scalloped trailing-edge extension produced the same lift at lower flapping frequency, though that study measured lift and power, not acoustics.

A fringe is set by tooth height $h$ and pitch $\lambda$, non-dimensionalised following~\cite{ji_influences_2022} as $k^* = 2\pi\bar c/\lambda$ and $h^* = h/\bar c$ at the mean chord $\bar c = 30\,\mathrm{mm}$. Five different fringe designs and one serrated version are manufactured using the baseline wing's frame layout, with acoustics and lift-force testing detailed in Section~\ref{sec:fringes}.

\subsection{Wing Venation}
The second mitigation strategy investigated in this paper focuses on minimising the noise generated during stroke reversal by optimised placement of wing venation patterns.
The resulting click at the end of each stroke is interpreted as an inertia-driven structural snap-through of the membrane, on three grounds. First, the forcing peaks at reversal. A material point at spanwise station $y$ sees transverse acceleration $\ddot\varphi\,y$, and at reversal the angular velocity vanishes while $|\ddot\varphi| = \Phi\omega^2$ is maximal, so the inertial load is largest exactly when the wing turns around, and the saturated pitch flip (Section~\ref{sec:kinematics}) concentrates almost all of the feathering acceleration at the same instant. An independent finite-element study places the inertial-load maximum at each reversal and finds the rapid rotation there amplifies inertial deformation roughly sixfold~\cite{hou_comparative_2022}. Second, an untensioned sheet has little restoring stiffness against this load. A membrane resists transverse loading through tension acting over curvature, and without pretension the inertial forcing can drive a dynamic inversion, whereas a static push on the same sheet merely stiffens it smoothly (Section~\ref{sec:dynamic}). The reversal is therefore a dynamic overshoot rather than a quasi-static buckling. Third, whether, where and how violently the sheet inverts is governed by the local bending rigidity $D$ and its distribution, which can be adjusted using bonded carbon strips. A well-positioned stiffener spanning the curl that tries to form can suppress the local inversion, whereas a misplaced one may even shift the instability to a less favourable location. The snap is thus layout-sensitive; hence, finding the most appropriate venation can be formulated as an optimisation problem. 

\section{Acoustic Optimisation Model}\label{sec:venation-model}
\begin{figure}[tb]
  \centering
  \definecolor{cOpt}{HTML}{2F6FB2}
  \definecolor{cSim}{HTML}{2C8C7C}

  \resizebox{0.9\linewidth}{!}{
  \begin{tikzpicture}[
    >={Stealth[length=3mm, width=2mm]},
    font=\small,
    base/.style={rounded corners=3pt, inner ysep=8pt, inner xsep=8pt,
                 minimum height=14mm, line width=0.8pt},
    sim/.style={base, draw=cSim, fill=white},
    opt/.style={base, draw=cOpt, fill=white},
    flow/.style={->, line width=1.5pt, draw=black!70},
    ctrl/.style={->, line width=1.5pt, draw=cOpt},
    note/.style={font=\small\itshape, text=black!70, fill=white, inner sep=2pt},
  ]

  % 1. Preparation (Top Left)
  \node[opt] (decode) {%
    \vbox{\hsize=65mm
      \centering
      \parbox{60mm}{\centering\textbf{Preparation}\\[8pt] Model receives and processes the design candidate geometry, meshed to 0.5\,mm}\\
      \vphantom{M}\\[4mm]
      \includegraphics[width=65mm,keepaspectratio]{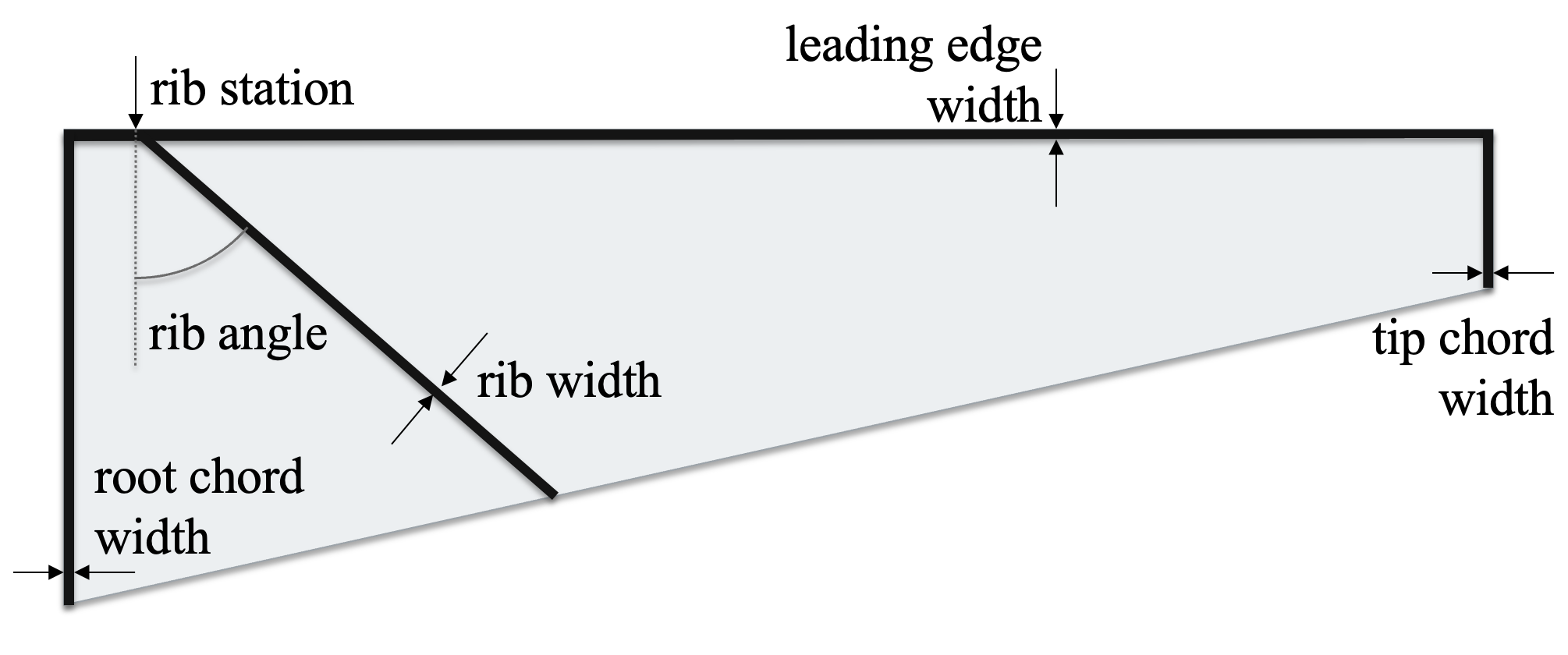}
    }
  };

  % 2. Solve (Top Right)
  \node[sim, right=15mm of decode] (solve) {%
    \vbox{\hsize=65mm
      \centering
      \parbox{60mm}{\centering\textbf{Solve}\\[8pt] Model solves the dynamic response in time and calculates the deflection of the wing}\\
      \vphantom{M}\\[4mm]
      \includegraphics[width=65mm,keepaspectratio]{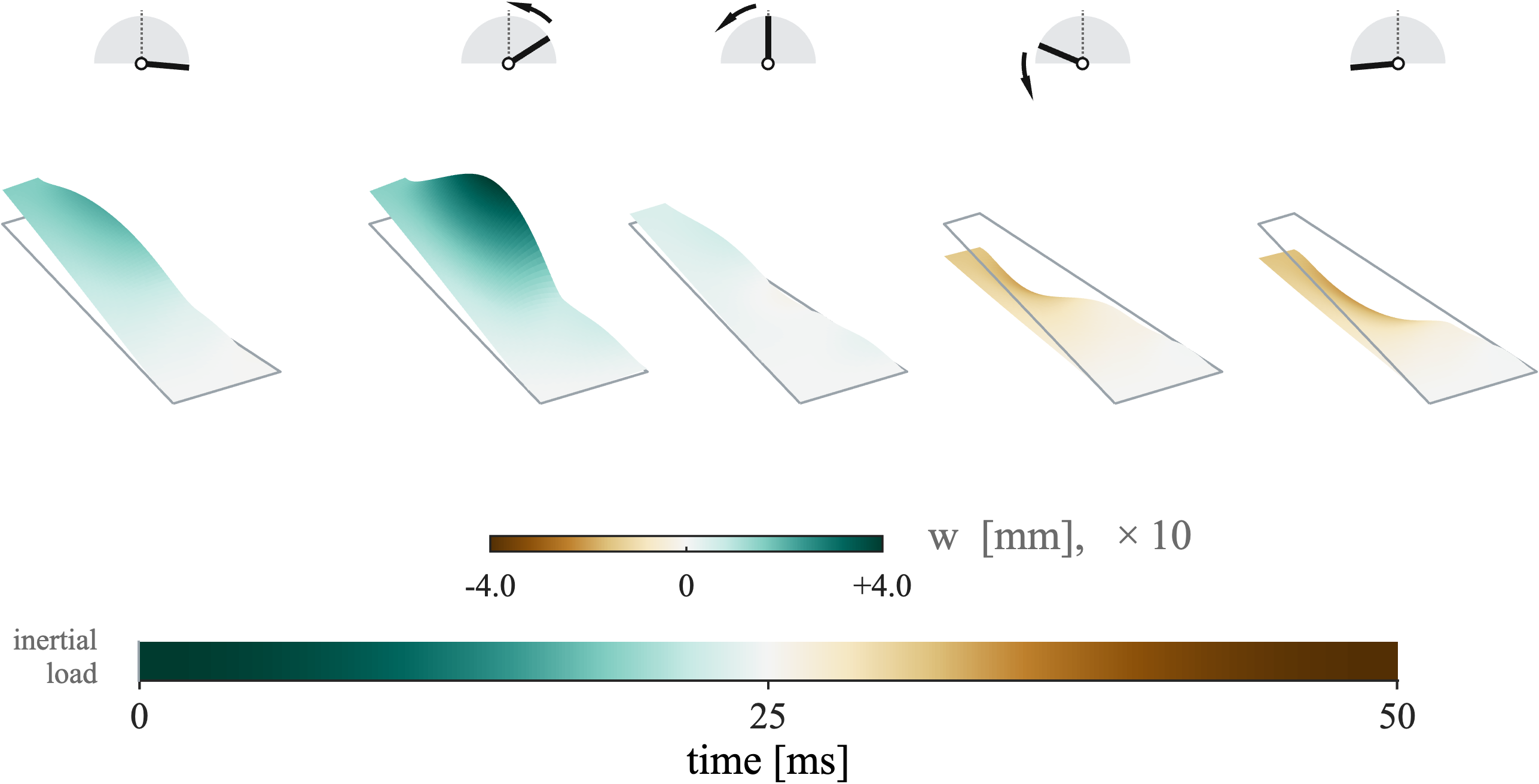}
    }
  };

  % 3. Score (Bottom Right)
  \node[sim, below=15mm of solve] (score) {%
    \vbox{\hsize=65mm
      \centering
      \parbox{60mm}{\centering\textbf{Objective evaluation}\\[8pt] High-passed rate of force on the air from the wing}\\
      \vphantom{M}\\[4mm]
      \includegraphics[width=65mm,keepaspectratio]{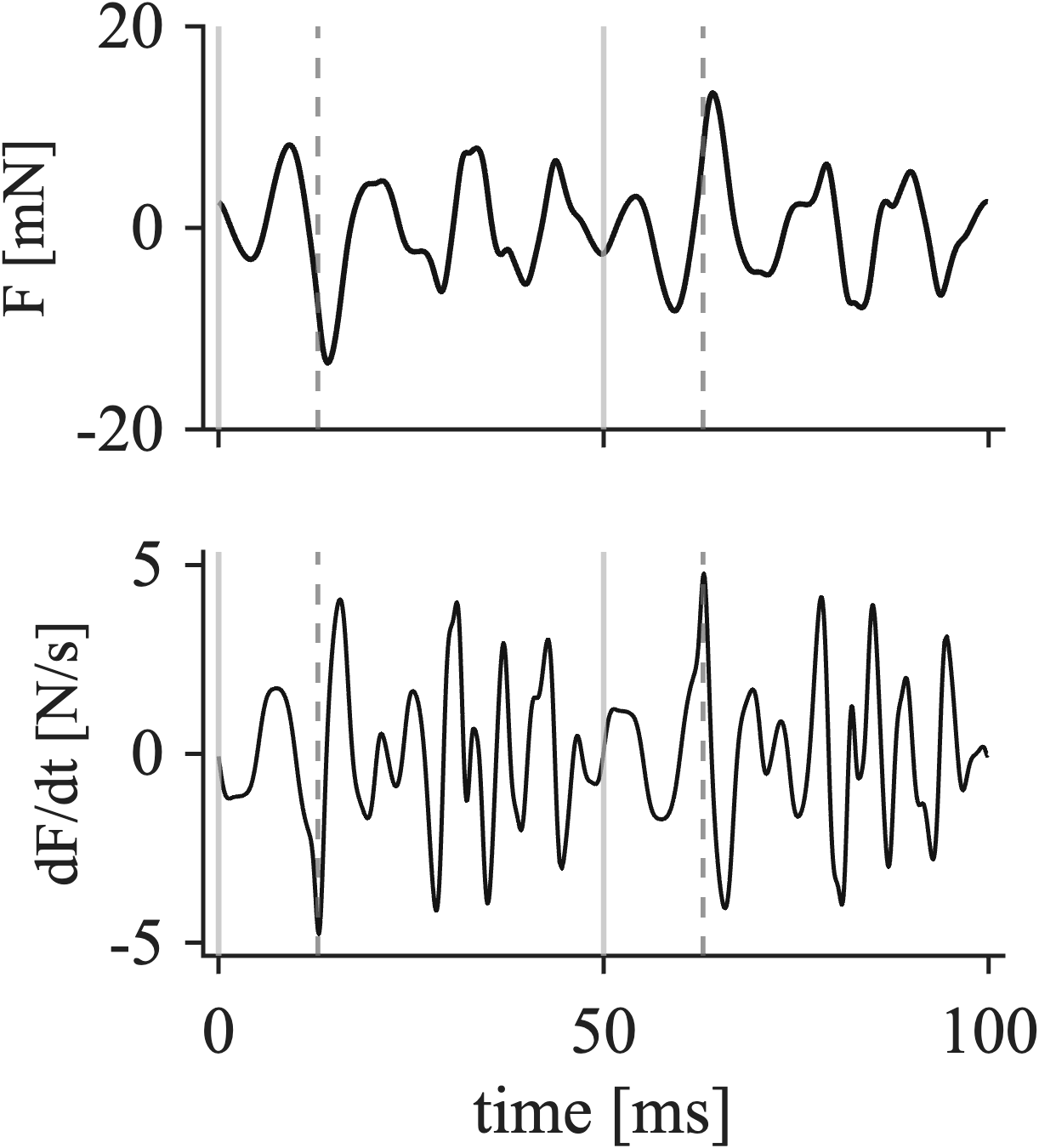}
    }
  };

  % 4. Search (Bottom Left)
  \node[opt, left=15mm of score] (search) {%
    \vbox{\hsize=65mm
      \centering
      \parbox{60mm}{\centering\textbf{Bayesian update}\\[8pt] Gaussian process, expected improvement picks the next design}\\
      \vphantom{M}\\[4mm]
      \includegraphics[width=65mm,keepaspectratio]{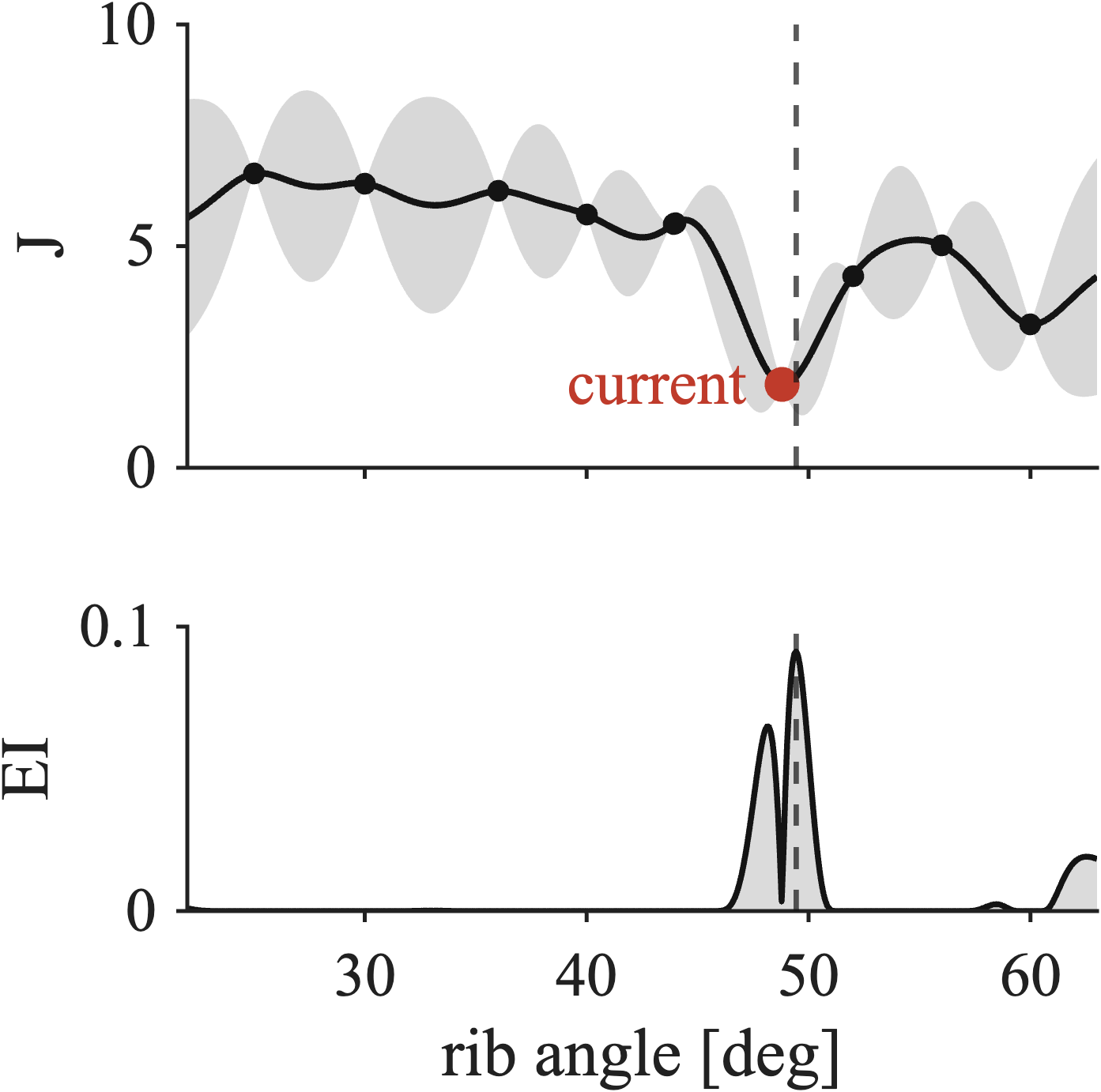}
    }
  };

  % Connecting Arrows forming the cycle
  \draw[flow] (decode.east) -- node[note] {$D(x,y)$} (solve.west);
  \draw[flow] (solve.south) -- node[note] {$w(x,y,t)$} (score.north);
  \draw[flow] (score.west)  -- node[note] {$J$} (search.east);

  % Return Path
  \draw[ctrl] (search.north) -- node[note, text=cOpt] {next design} (decode.south);

  \end{tikzpicture}
}% End of resizebox
\caption{An overview of the iterative process of the optimisation model.}
\label{fig:pipeline}
\end{figure}

This section develops a method for choosing a quiet venation layout on a flexible membrane wing. Three terms are used throughout. The frame is the outer part of that plate, the leading edge, the root and tip chord cantilevered from the drive pin. The venation is the part of the same plate that extends from the frame across the membrane, and a rib is one such extension, defined by the spanwise station where the rib extends from the leading edge, the rib width and the rib angle with respect to the chord. The inputs to the optimisation model are the planform, the membrane areal mass $\rho_m h_m$ and in-plane stiffness $E_m h_m$, the rib modulus $E_{CF}$ with the available plate thicknesses $h_{CF}$, the prescribed kinematics, and a mass budget for the total stiffening material, frame and venation. The output of the model is the frame widths and the venation of the quietest wing found. The optimum is found by repeating the loop of Fig.~\ref{fig:pipeline}. A candidate layout is converted into a wing geometry, the plate model is solved in time for that wing through stroke reversal to give the deflection of the membrane at every point and instant, an objective function turns that deflection history into a measure of the click strength, and a Gaussian process fitted to every objective value so far selects the next layout.

\label{sec:kinematics}
The stroke is $\varphi(t) = \Phi\cos\omega t$, with $\Phi$ the stroke amplitude, $\omega$ the angular frequency and $f$ the flapping frequency. The feathering rotation, i.e. the passive pitching of the wing about its leading edge, is not sinusoidal, because a leading edge free to rotate between mechanical stops rests against a stop through most of the half-stroke and turns over quickly near stroke reversal. The pitch angle $\theta(t)$ is therefore represented by a saturating sinusoid:

\begingroup\small
\begin{equation}
    \theta(t) = \Theta\,\frac{\tanh\!\big[k\cos(\omega t + \psi)\big]}{\tanh k}
    \label{eq:pitch}
\end{equation}
\endgroup
Here, $\Theta$ is the stop angle, $\psi$ is the phase of the flip relative to the stroke, and $k$ is a parameter used to shape the wave. This formulation concentrates almost all of the pitch angular acceleration into the flip just after stroke reversal, instead of distributing it across the half-stroke. A user of the method supplies their own amplitudes, frequency and stop angle to the same equation. The wing used here runs at $\Phi = 95^\circ$ and $f = 10\,\mathrm{Hz}$ against stops at $\Theta = 40^\circ$, with $\psi = -2.17\,\mathrm{rad}$ and $k = 3.0$.

\subsection{Structural model}\label{sec:dynamic}
The wing is modelled as a von K\'arm\'an plate, a thin sheet whose out-of-plane deflection $w$ may be large relative to its thickness while material strains remain small. The out-of-plane equilibrium of a von K\'arm\'an plate gives the following balance equation:

\begingroup\small
\begin{equation}
    \underbrace{D\,\nabla^4 w}_{\text{bending}}
    - \underbrace{\big(N_x w_{xx} + 2N_{xy}w_{xy} + N_y w_{yy}\big)}_{\text{tension through curvature}}
    = p 
    \label{eq:vk}
\end{equation}
\endgroup
Here, $D$ is the local bending rigidity, $N_x$, $N_y$ and $N_{xy}$ the in-plane membrane forces, $w_{xx}$, $w_{yy}$ and $w_{xy}$ the curvatures, and $p$ the transverse load. Transverse loads lead to both bending and membrane tension acting through curvature. The venation layout determines which of these effects dominates. The proposed model accounts for the membrane in terms of its areal mass $\rho_m h_m$ and its in-plane stiffness $E_m h_m$. Given the material properties of the bare membrane in Section~\ref{sec:framework}, the bending rigidity of the membrane is the following:

\begingroup\small
\begin{equation}
    D_m = E_m h_m^3/[12(1-\nu_m^2)] = 1.70\times10^{-4}\,\mathrm{N\,m}
\end{equation}
\endgroup
Therefore, the bare sheet behaves as a near-membrane, and almost all of its resistance to transverse loads comes from its tension, acting through curvature.

Differentiating the potential energy of the discretised plate twice with respect to the nodal displacements $\mathbf{d}$ gives the consistent tangent $\mathbf K_T = \mathbf K_{mat} + \mathbf K_{uw} + \mathbf K_g$. The terms in the expression represent, respectively, the material stiffness, the von K\'arm\'an coupling between the in-plane displacements $u$ and $w$, and the geometric stiffness of the membrane force acting through the curved shape. As the sheet bends, $\mathbf K_g$ can drive $\mathbf K_T$ to lose positive definiteness: this is where the reversal snap occurs. Earlier venation studies built on linear or static formulations carry no geometric stiffness term $\mathbf{K_g}$~\cite{stanford_aeroelastic_2012,reade_investigation_2022,zhu_parametric_2025}. The von K\'arm\'an plate model of Eq.~\ref{eq:vk} is considered to hold~\cite{hakim_innovative_2023,ghassemi_comparison_2018} while the deflection-to-thickness ratio stays moderate, $w/h_m \lesssim 5$--$10$, so every deflection is reported with its $w/h_m$.

The tangent describes how the plate resists deflection at any instant, while the reversal snap is driven by the inertia of the moving wing, so the plate must be followed through the stroke. Adding inertia and damping to the internal force gives the equation of motion of the discretised plate,

\begingroup\small
\begin{equation}
    \mathbf M\ddot{\mathbf d} + \mathbf C\dot{\mathbf d} + \mathbf f_{int}(\mathbf d) = \mathbf f_{ext}(t)
    \label{eq:eom}
\end{equation}
\endgroup
in the nodal displacements $\mathbf d$. Here $\mathbf M$ is the mass matrix, $\mathbf C$ the damping matrix, $\mathbf f_{int}$ the nonlinear internal force, and $\mathbf f_{ext}$ the applied transverse load of Eq.~\ref{eq:forcing}. No stiffness matrix appears in Eq.~\eqref{eq:eom}, because the restoring force of the plate is nonlinear in $\mathbf d$ and is carried by $\mathbf f_{int}$. Eq.~\eqref{eq:eom} is integrated in time by the implicit generalised-$\alpha$ method at spectral radius $\rho_\infty = 0.8$~\cite{chung_time_1993}, which damps unresolved high-frequency modes of the ribs and the frame without smearing the reversal event, and the nonlinear equations at each time step are solved by Newton's method on $\mathbf K_T$ to a relative residual of $10^{-6}$.
The membrane loses energy as it flexes, and this loss is represented by the stiff-proportional damping, $\mathbf C = \beta_K\mathbf K_{lin}$, with $\mathbf K_{lin}$ the linear stiffness matrix. The damping force $\beta_K\mathbf K_{lin}\dot{\mathbf d}$ is the elastic force multiplied by $\beta_K$ and by the rate of deformation, so for a given amplitude, a fast vibration, whose rate of deformation is high, meets a far larger damping force than a slow one. In modal terms, each vibration mode receives a loss factor equal to $\beta_K$ times the angular frequency of that mode, so fast, small-scale vibration is damped more strongly than the slow bending of the wing as a whole. The coefficient $\beta_K = 5\times10^{-5}\,\mathrm{s}$ is set so that the model loses as much energy per flexing cycle as the material itself does, matched at the centre of the analysis band~\cite{rodriguez-perez_dynamic_2000}. 

A check was performed to ensure that the reversal snap is a dynamic event. Instead of the inertial load, the wing was pushed with a slowly increasing static force. Using the arc-length continuation~\cite{crisfield_fast_1981} method, this solver can follow the load-deflection path through a snap if there is one. The check showed that no snap was found: the membrane just became stiffer as the static force increased. This shows that the inversion at stroke reversal needs the inertia of the moving wing.

The transverse load is the inertia of the wing reacting against the prescribed rigid-body rotations:

\begingroup\small
\begin{equation}
    \mathbf f_{ext}(t) = -\ddot\varphi(t)\!\int_S \mu\,y\,\mathbf N_w\,\mathrm dS
                        \;-\;\ddot\theta(t)\!\int_S \mu\,(x-x_a)\,\mathbf N_w\,\mathrm dS
    \label{eq:forcing}
\end{equation}
\endgroup
Here, $S$ is the plate mid-surface, $\mathrm dS$ its area element, $\mathbf N_w$ the out-of-plane shape functions, $\mu(x,y)$ the local areal mass of membrane and venation, $y$ the spanwise distance from the root, $x$ the chordwise coordinate and $x_a$ the chordwise position of the pitch axis.

The air entrained by the deflecting surface is carried as an added mass $\mu_{add}(y) = \tfrac{\pi}{4}\rho_{air}c(y)$, with $\rho_{air}$ the air density and $c(y)$ the local chord. On our tapered wing, the added mass is not a constant, but decreases from $43.3$ to $14.4\,\mathrm{g/m^2}$ between the wing root and tip. Aerodynamic pressure is not modelled because the load at stroke reversal is predominantly inertial, and accounting for unsteady aerodynamic effects would lead to prohibitive computational cost.
Therefore, the entrained air is retained as inertia~\cite{combes_into_2003,hou_comparative_2022}. In-plane centrifugal tension, a few per cent of the loading and identically zero at stroke reversal, is likewise omitted. Eq.~\eqref{eq:forcing} also sets the economics of the problem, since forcing grows with mass weighted by distance from the root, which makes structural mass self-penalising.

Venation is represented in the model through the stiffness terms $D$ and $Eh$ of Eq.~\ref{eq:vk}, which vary over the wing.
Wing layouts are expressed as an occupancy field $\chi(x,y)\in\{0,1\}$, equal to one where a rib covers the membrane and to zero where the membrane is bare. The plate stiffness becomes $D(x,y) = D_m + \chi\,D_{CF}$ and $(Eh)(x,y) = E_m h_m + \chi\,E_{CF} h_{CF}$, where $D_{CF}$ is the bending rigidity of by a rib and $E_{CF} h_{CF}$ is the in-plane stiffness of a rib of thickness $h_{CF}$. A rib bonded to one face of the membrane contributes its own bending stiffness $D_{CF} = E_{CF} h_{CF}^3/{12}$.

\subsection{The loading-rate surrogate, design space and optimisation}
The optimisation requires a computationally cheap click-strength measure, so no acoustic propagation runs inside the search. Curle's result~\cite{curle_1955} gives the far-field pressure radiated by a body that is small compared to the acoustic wavelength~\cite{ffowcs_williams_sound_1969}, as a dipole whose strength is the rate of change of the net force the body exerts on the air,

\begingroup\small
\begin{equation}
    p'(\mathbf x,t) \approx \frac{x_i}{4\pi c_0 r^2}\,\frac{\partial F_i}{\partial t}\Big(t-\frac{r}{c_0}\Big),
    \label{eq:curle}
\end{equation}
\endgroup

where $p'$ is the acoustic pressure at the observer position $\mathbf x$ with components $x_i$, $r$ the distance from the body, $c_0$ the speed of sound and $F_i$ the net force on the air. 
The point-dipole form of Eq.~\ref{eq:curle} holds for sound frequencies below $f_{compact} = c_0/(2\pi L)$, about 400\,Hz for the wing span $L = 135\,\mathrm{mm}$, which covers the 80--300\,Hz range in which the reversal click radiates~\cite{wu_experimental_2026}.

The model does not consider any aerodynamics, so the only force the wing exerts on the air is the reaction of the entrained air mass,

\begingroup
\small
\begin{equation}
    F(t) = \int_S \mu_{add}\,\ddot w\,\mathrm dS
    \label{eq:force}
\end{equation}
\endgroup

which is the net force $F_i$ of Eq.~\ref{eq:curle} in the direction normal to the wing. 
Using Eq.~\ref{eq:curle}, the far-field pressure is proportional to $\mathrm{dF}/\mathrm{dt}$, so the rate of change of force the wing exerts on the air is taken as the measure of the radiated click, and the objective the search minimises is the high-passed rate of the force in Eq.~\ref{eq:force},

\begingroup\small
\begin{equation}
    J = \mathrm{rms}\Big[\,\mathrm{highpass}_{100\,\mathrm{Hz}}\big(\mathrm dF/\mathrm dt\big)\Big]
    \label{eq:objective}
\end{equation}
\endgroup
The peak in $\mathrm dF/\mathrm dt$ occurs at the wing-pitch flip just after stroke reversal. The $100\,\mathrm{Hz}$ high-pass, implemented as a fourth-order zero-phase Butterworth filter, removes the wingbeat frequency and its low harmonics, which grow with stiffness and do not correspond to the click. 
The root mean square (rms) is taken over three flap cycles, simulated at 960 steps per cycle. In each simulation, the first flap cycle is discarded as a start-from-rest generates transients several times larger than the periodic response.

Each layout is defined by 47 parameters. Six are the start and end widths of the leading edge, the root and tip chord, two per piece. One is the plate thickness shared by the frame and the ribs. The remaining forty belong to ten rib slots, each carrying a presence flag, a station, a width and an angle. The parameters are converted deterministically to a buildable wing, and the conversion enforces manufacturability. The wing is cut as one connected piece, so disconnected, branching or curved layouts are excluded. Frame widths fold to a monotonically narrowing taper, since all three members cantilever from the root drive pin. Ribs within a $2\,\mathrm{mm}$ gap at the leading edge are merged, and no frame piece or rib is narrower than 0.8\,mm, the machining limit of our in-house water-jet.
The rib angle is free to take any value up to $\pm60^\circ$ either side of the chordwise direction.

Mass is the only hard constraint, computed in closed form before any optimisation run, so a design heavier than the limit is rejected without simulation. The allowable mass budget is 400\,mg of total carbon fibre. The search accepted designs up to 490\,mg so that the trend of the objective with mass could be observed on both sides of the budget, but only designs within the budget were eligible for manufacture.

The objective, $J$, in Eq.~\ref{eq:objective} is sensitive to the fast vibrations that the damping term in Eq.~\ref{eq:eom} controls. Since $\beta_K$ is chosen rather than measured, the actual physical effect of damping on the click cannot be quantified by this model. Instead, we checked whether different values of $\beta_K$ change the optimisation result by doubling and halving the selected value of $\beta_K$ and rerunning the model. In both cases, the objective value $J$ for every design moved in the same direction, and the relative ranking of the designs did not change. Therefore, the model can be used to compare designs against each other, albeit not to claim an absolute noise level for each individual design. 

The optimisation ran 40 initialisation designs, then a Bayesian tail to 180 evaluations, retaining every force and deflection history for re-analysis. All the evaluations in the model reached a converged solution. Across the 180 designs, the objective spans $J=1.89$--$10.30$, carbon-fibre mass spans from 123--490\,mg, and peak deflection-to-thickness $w/h_m$ from 3.4 to 7.9.

Three trends are observed from these evaluations:
Firstly, the cost rises with the spanwise centroid of the venation. Designs with venation near the root give a lower cost than designs with venation near the tip.
The designs yielding the lowest cost have a venation centroid within 20\,mm of the root. This can be explained by the forcing of Eq.~\ref{eq:forcing}. Each element of mass is loaded in proportion to the distance of that mass from the root, so a rib placed far outboard is thrown harder at every stroke reversal than the same rib placed near the root, and the extra inertial load outweighs the stiffening benefits the rib has. 
Secondly, the best designs place the ribs at an angle, instead of vertically down the chord. All twenty best designs placed the rib at an angle beyond $15^\circ$ from the chord. A possible explanation is that during stroke reversal the wing twists about the leading edge, and a twisted wing deforms in shear rather than in bending along the chord. Hence, a diagonal rib is effective as it crosses the principal directions of the twist and stiffens the wing. 
Lastly, the mass budget is not an active constraint. The optimum design sit at 310\,mg, far from the imposed limit of 400\,mg. As seen in Eq.~\ref{eq:forcing}, every gram of carbon fibre adds to the inertial load at stroke reversal. Extra material will only lower the cost when the stiffness that it provides outweighs the load that it adds. Adding carbon fibre anywhere else makes the wing louder; therefore, the search stops adding well before the constraint.

The optimum design, M1, features a single diagonal rib at $48.8^\circ$, 0.92\,mm wide. The objective value of M1 is $J=1.89$.
Having a single-rib optimum makes rib orientation testable in isolation. The adverse twin design, M1-adverse, has the same frame, rib station and rib width, but differs in rib angle, with its rib placed at $43.9^\circ$. The model calculated that this minor change in rib orientation increases the cost to $J=5.50$, a factor of 2.9 when compared to M1. Against the best bare frame, M1 reduces the cost from $J = 4.77$ to $J = 1.89$, a factor of 2.5, for 52\,mg more carbon-fibre material.

\section{Experimental Evaluation}\label{sec:venation-results}
The model of Sec.~\ref{sec:venation-model} ranks the designs by a surrogate, so the experiments in this section test whether the predicted ranking is accurate. Selected venation designs and wings with trailing-edge fringes are manufactured and tested for their acoustic output against the same baseline wing at the same operating point.

\subsection{Experimental Setup}
Every wing tested here is driven by the passive pitch flapping mechanism by Broers and Armanini \cite{broers_multimodal_2026}, shown in Fig.~\ref{fig:jig}. The flapping frequency, $f_0$, is set to 10\,Hz for all tests to ensure a fair comparison, as the radiated sound depends on the kinematics. Motor current is adjusted for each wing to reach that frequency, because different wing geometries change the wing's inertia, so each wing draws a slightly different current for the same motion.

\begin{figure}[htb]
    \centering
    \includegraphics[width=\columnwidth]{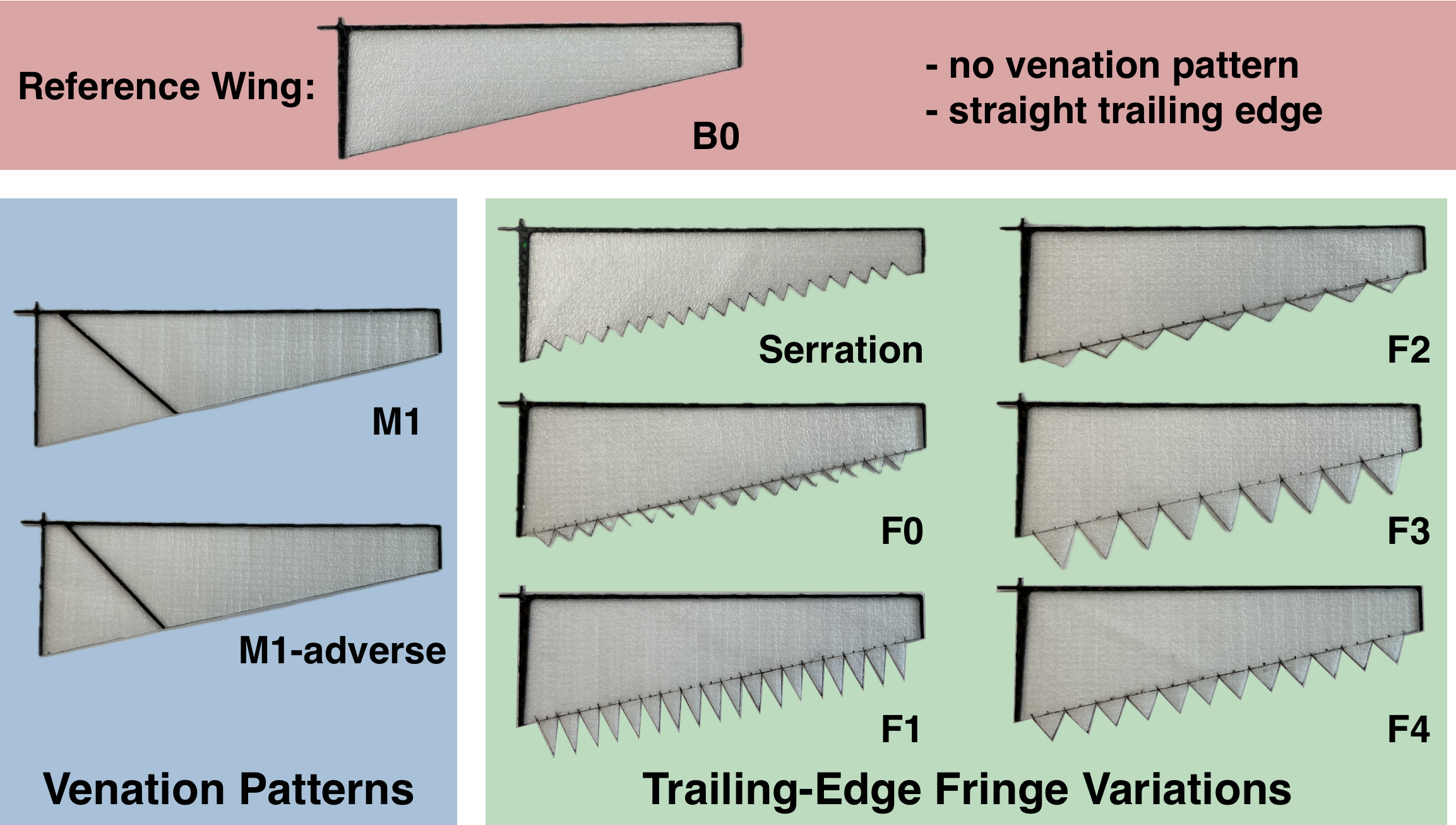}
    \caption{Manufactured wing models with the reference wing, the venation pattern set, and the edge-treatment set.}
    \label{fig:buildset}
\end{figure}

Two sets of wings were manufactured and compared to the baseline wing B0. The venation pattern set comprises two rib-placements (M1, M1-adverse) selected by the model of Sec.~\ref{sec:venation-model}.
The edge-treatment set has five additive fringes of different geometries (F0--F4) and one subtractive geometry with triangular serrations. The individual geometries are given in Table.~\ref{tab:fringe}. All wings share the same manufacturing process described in Sec.~\ref{sec:framework}, and are shown in Fig.~\ref{fig:buildset}.

Sound is recorded with a miniDSP UMIK-1 USB measurement microphone, sampled at 48\,kHz. Each recording is stored as raw audio and analysed offline. Three equidistant microphone positions are used, all on a horizontal semicircle of radius 500\,mm about the flapping axis, on the same plane as the wing. At mid-stroke, the wing points at the $90^\circ$ position. Each wing is recorded twice at each position. While the microphone cannot record absolute SPL levels, this experimental setup allows us to compare different wing recordings made on the same day at the same position and distance. Each wing is therefore compared with a baseline wing built and recorded on the same day, under the same conditions and at a matched frequency $f_0$.

The recorded sound difference at each stroke reversal is read from a band-limited envelope. The recording is first band-passed with a fourth-order zero-phase Butterworth filter. The envelope is the magnitude of the Hilbert transform of that band-limited signal, which traces the peaks of the oscillating trace, and is then smoothed over about 1\,ms. For venation, the band runs from 80\,Hz to 1000\,Hz. The lower edge of the band removes the 10\,Hz flapping tone and low harmonics, and the upper edge of the band sits just below the noise of the flapping mechanism, which is common to every wing and would drown out any wing effect. Wu et al.~\cite{wu_experimental_2026} place the impulsive loading content of a comparable wing in this range, with the click at stroke reversal concentrated at 80--300\,Hz and the trailing edge vortex shedding at 300--1000\,Hz, so the trailing-edge fringes of Sec.~\ref{sec:fringes} are evaluated in the narrower 300--1000\,Hz band.

Stroke reversals are located as peaks of the band envelope spaced half a flapping period apart, which gives two per cycle. A window of ±15\,ms is taken about each detected reversal and pooled across the recordings of that wing, giving between 300 and 400 windows per wing and position. The amplitude reported for a wing is the median, over its windows, of the largest envelope value inside each window. The reported reduction for wing $i$ can be written as the following equation:

\begingroup\small
\begin{equation}
    \Delta L_i = 20\log_{10}\big(\hat A_i / \hat A_{\mathrm{B0}}\big)
    \label{eq:dl}
\end{equation}
\endgroup
Here, $\hat A_i$ represents the median per-reversal peak envelope amplitude of wing $i$ and $\hat A_{\mathrm{B0}}$ that of the same-day baseline, and a negative $\Delta L$ is an acoustic output reduction when compared to the baseline wing B0.
Additionally, the lift force generated by each wing is measured using an ATI Nano 17Ti six-axis force/torque sensor, recording at 1000\,Hz. This setup includes a Hall-effect position sensor which records the exact position of the wing's leading edge.
Each recording session is run for at least 7\,s and the recorded lift forces are cycle-averaged using the wing position data.

\subsection{Wing Venation Testing}
Three wings are assessed: the bare-frame baseline B0, M1, and M1-adverse (Fig.~\ref{fig:buildset}). B0 produced a cycle-mean lift of 0.102\,N, M1 a lift of 0.112\,N. Therefore, optimal venation design produces comparable lift compared to the baseline.

Table~\ref{tab:venation-results} compares each wing against its baseline. As can be seen, the overall A-weighted level barely moves, with every venated wing within 0.8\,dB of the frame, because A-weighting and the flapping mechanism's noise above 1\,kHz dominate the broadband level and a change confined to the low-frequency snap does not register. However, within the 80--1000\,Hz snap band, the noise reduction is large and consistent. At the $90^\circ$ station, M1 falls 7.8\,dB below the frame, while the mechanism floor above 1\,kHz is unchanged to within ±0.8\,dB, confirming the reduction is due to the wing.

\begin{table}[htb]
\centering
\caption{Experimental Results: Venation Optimisation}
\label{tab:venation-results}
\begin{tabularx}{\columnwidth}{@{}l *{5}{>{\centering\arraybackslash}X}@{}}
\hline
 & $\Delta$dBA & \multicolumn{3}{c}{$\Delta L$ (dB)} & Lift \\
Wing & overall & $0^\circ$ & $90^\circ$ & $180^\circ$ & (N) \\
\hline
B0         & ref    & ref    & ref    & ref    & 0.102 \\
M1         & $-0.8$ & $-2.3$ & $-7.8$ & $-4.1$ & 0.112 \\
M1-adverse & $-0.7$ & $+0.9$ & $-0.2$ & $-1.0$ & n/a   \\
\hline
\multicolumn{6}{@{}p{\columnwidth}@{}}{A negative value is a reduction. $\Delta L$ is the peak reduction in the $80$--$1000\,\mathrm{Hz}$ band at each microphone position, and lift is the cycle-averaged vertical force.}
\end{tabularx}
\end{table}

\begin{figure}[htb]
\centering
\includegraphics[width=0.8\columnwidth]{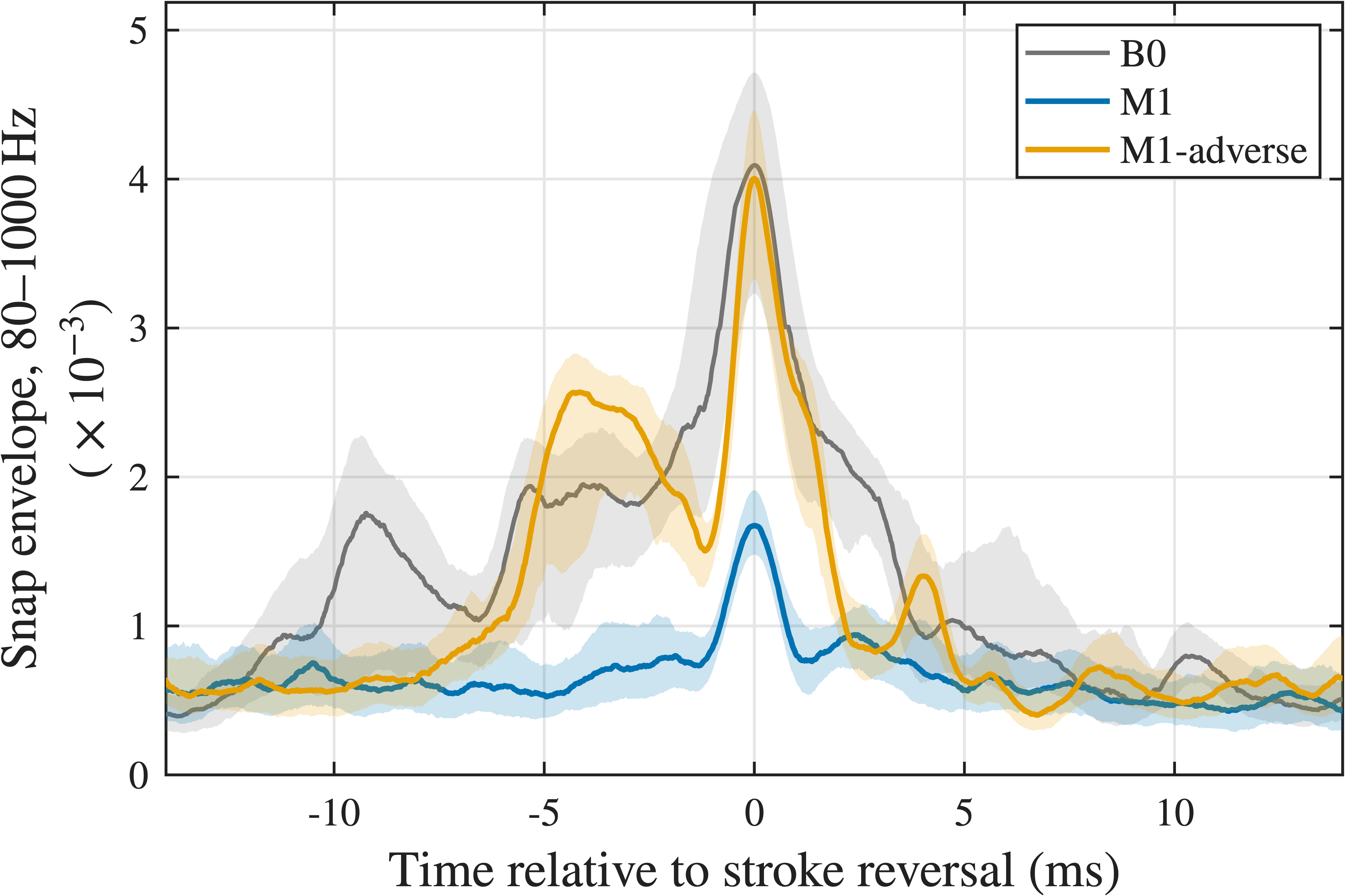}
\caption{Reversal-averaged snap envelope at the $90^\circ$ microphone position for baseline wing B0, venation set M1 and M1-adverse, in normalised amplitude.}
\label{fig:acoustic}
\end{figure}

M1 and M1-adverse are identical except for a 5$^{\circ}$ difference in rib orientation angle. Fig.~\ref{fig:acoustic} shows the three wings' acoustic output at the $90^\circ$ position on the same axes. The shaded band marks the 25th to 75th percentile over reversals. The click appears as the peak at zero, and the peak level is the quantity behind $\Delta L$ from Eq.~\ref{eq:dl}. M1 reduces the click by 7.8\,dB against the same-day B0 frame, while M1-adverse stays within a decibel of B0 at all three positions. M1 reduces the click by 7.8\,dB at the $90^\circ$ position against the same-day B0 frame (Fig.~\ref{fig:acoustic}), while M1-adverse stays within a decibel of B0 at all three positions. Compared against each other, M1 and M1-adverse differ by 7.6\,dB at $90^\circ$ and by about 3\,dB at the other two positions.

\subsection{Bioinspired Trailing-Edge Fringes Testing}\label{sec:fringes}
Five additive fringes and one serrated wing were built (Fig.~\ref{fig:buildset}). Four of the fringes (F0--F3) combine a short or a tall tooth with a fine or a coarse pitch, allowing the effect of tooth height and pitch to be tested. The fifth wing (F4) sits at the middle of both ranges. The serrated wing features a triangular serration with a short-fine tooth geometry. The detailed geometries are given in Table.~\ref{tab:fringe}. 

Measured at the same flapping frequency, F0 produced a cycle-mean lift of 0.113\,N, and the cut serration version of the wing with the same tooth geometry produced 0.092\,N. The additive form thus generates about 25\% more lift for the same tooth geometry. Compared to B0, with 0.102\,N lift produced, F0 is higher, while F2 and F4 generate about the same lift (0.100\,N). F3 generates lower lift at 0.091\,N, while the tall and fine fringe of F1 generate the lowest lift at 0.077\,N. 

\begin{figure}
\centering
\includegraphics[width=0.8\columnwidth]{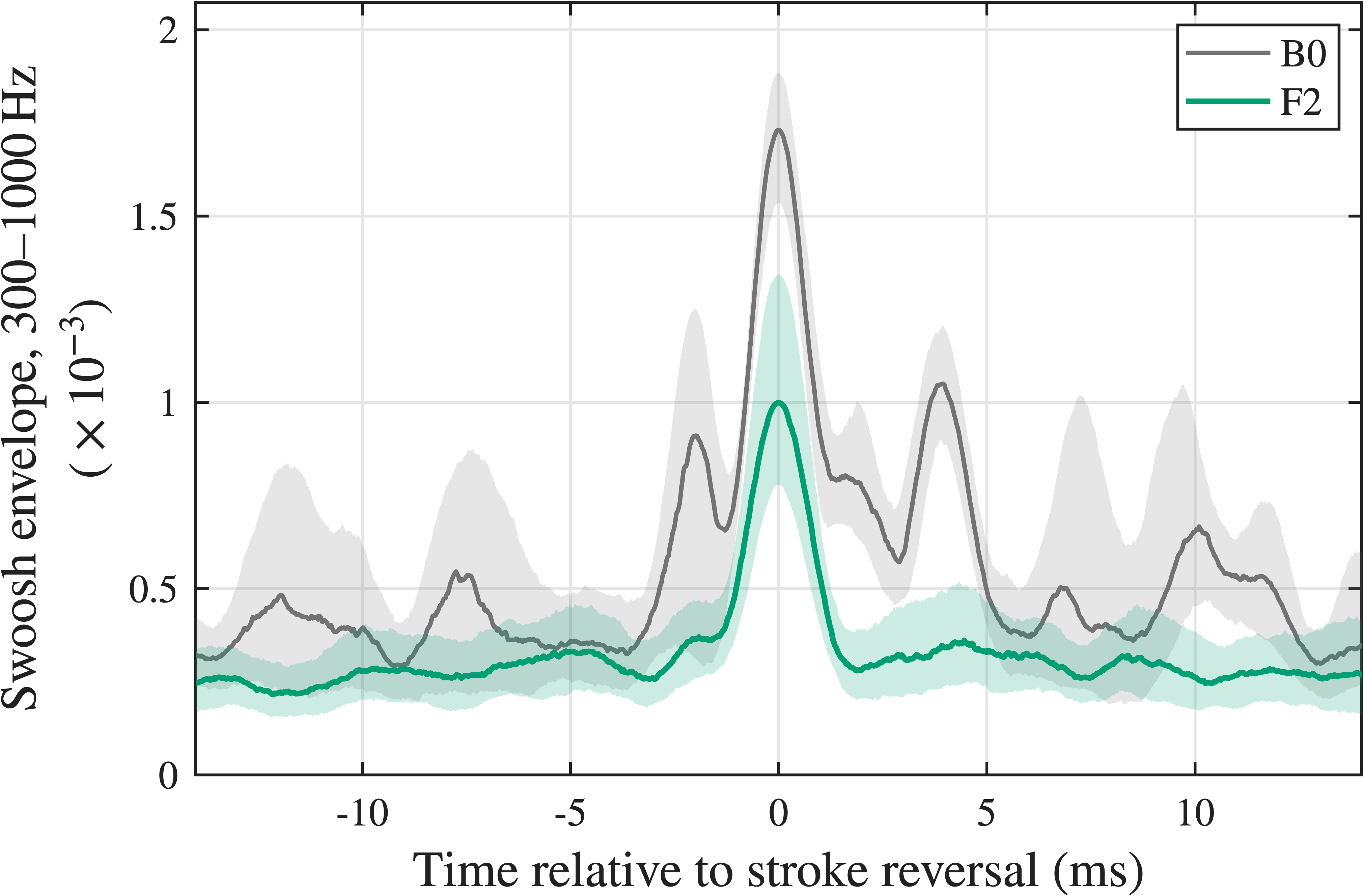}
\caption{Reversal-averaged swoosh envelope at the $90^\circ$ microphone position for the baseline wing B0 and the fringe pattern wing F2 in normalised amplitude.}
\label{fig:fringe-h2h}
\end{figure}

Table~\ref{tab:fringe} gives the 300--1000\,Hz noise levels compared to B0. As seen, the short teeth give a larger noise reduction. Averaged over the three microphone positions, the short fringes reduce the acoustics by 3.1\,dB for F0 and 4.1\,dB for F2, while the tall fringes reduce the acoustics by 1.1\,dB for F1. F3 did not show any improvement of acoustics suppression. Averaging over pitch, going from short to tall teeth adds 3.3\,dB, whereas averaging over height, doubling the pitch from 7\,mm to 14\,mm changes the level by 0.3\,dB. Height has a far more dominant impact than pitch. The tall geometries are also the ones that produced less lift, likely because slender teeth struggle to maintain their shape.
Fig.~\ref{fig:fringe-h2h} shows F2 outperforming the baseline wing B0.

\begin{table}[htb]
\centering
\caption{Experimental Results: Trailing-Edge Fringe Patterns}
\label{tab:fringe}
\begin{tabular}{lcccccc}
\hline
 & $h$ & $\lambda$ & \multicolumn{3}{c}{$\Delta L$ (dB)} & Lift \\
Wing & (mm) & (mm) &  $0^\circ$ & $90^\circ$ & $180^\circ$ & (N) \\
\hline
B0 (plain)   & --  & -- & ref    & ref    & ref  & 0.102 \\
F0 (short, fine)   & 5   & 7    & $-2.1$ & $-3.4$ & $-3.8$ & 0.113 \\
F1 (tall, fine)    & 13  & 7   & $+0.3$ & $-1.5$ & $-2.0$ & 0.077 \\
F2 (short, coarse) & 5   & 14    & $-2.1$ & $-4.8$ & $-5.3$ & 0.100 \\
F3 (tall, coarse)  & 13  & 14   & $+0.7$ & $+1.7$ & $-1.2$ & 0.091 \\
F4 (centre)        & 9   & 10.5 & $+1.1$ & $+1.4$ & $+1.1$ & 0.100 \\
Serration          & 5   & 7    & $-2.0$ & $-6.3$ & $-5.3$ & 0.092 \\
\hline
\multicolumn{7}{l}{\begin{tabular}[c]{@{}l@{}}Measured lift and acoustic reduction of trailing-edge fringe patterns against \\ the baseline wing B0. A negative value is a reduction. The cut triangular\\ serration carries the F0 geometry.\end{tabular}}
\end{tabular}
\end{table}

Averaged over the three positions, the cut serration reduces the band level by 4.5\,dB, and the fringe of matched tooth geometry F0 by 3.1\,dB. The best fringe wing, F2, reduces the band level by 4.1\,dB. At the $90^\circ$ position, the cut serration is 1.5\,dB better than F2, and at $0^\circ$ and $180^\circ$ the two are equal. The additive fringe therefore comes within about half a decibel of the cut serration on average, while holding about a quarter more lift at matched tooth geometry.

\section{Conclusion}\label{sec:conclusion}
This paper presented a wing design framework for reducing the acoustic signature of an FWMAV. Two approaches were proposed and tested to reduce the noise produced by a flapping wing.
Firstly, we investigated wing designs with an integrated venation bonded onto one side of the wing membrane, with the goal of increasing stiffness and reducing wing deformation during stroke reversal. Based on the assumption that loading rate is a surrogate for sound production, a nonlinear model was derived and applied within an optimisation setup to determine an optimal venation layout. 
The optimisation showed that rib positioning plays an essential role in sound production. The quietest layout, M1, features a single rib near the root, at a $48^\circ$ angle to the chord. Compared to the baseline wing, this venation layout reduced the click at stroke reversal by 7.8\,dB, within the band of 80--1000\,Hz. This configuration also generated more lift than the baseline, and only had a 50\,mg higher mass due to the added carbon-fibre material. Repositioning the rib by 5$^{\circ}$ with respect to the M1 positioning removed the noise suppression benefit again, proving that correct venation positioning is crucial.
The second proposed noise-reduction solution is an additive trailing-edge fringe, inspired by the trailing-edge feather pattern in extended hummingbird wings. This feature was found to reduce the broadband swoosh within the 300--1000\,Hz band, while maintaining wing area and thus producing sufficient lift. Experiments showed that short teeth, of height 5\,mm or $h^*=0.11$ of the mean chord, reduced the broadband noise by 3--4\,dB. The improvement was comparable to that achieved by a subtractive serrated wing, while generating 25\% more lift by not cutting into the wing membrane.
While significant noise reduction was achieved, the drive of the flapping mechanism still dominated the sound above 1\,kHz, limiting the overall noise reduction achievable. Future work includes investigating a quieter flapping mechanism and a wing featuring both the venation layout and the short additive fringe, to test the extent to which the two proposed solutions can work together to further suppress noise.

\section*{Acknowledgment}
We acknowledge the use of AI (Claude Opus 4.8) for code implementation and debugging purposes. Hummingbird picture by Pexels user Robert So.

\bibliographystyle{ieeetr}

\bibliography{references}

@ARTICLE{curle_1955,
       author = {{Curle}, N.},
        title = "{The Influence of Solid Boundaries upon Aerodynamic Sound}",
      journal = {Proceedings of the Royal Society of London Series A},
         year = 1955,
        month = sep,
       volume = {231},
       number = {1187},
        pages = {505-514},
          doi = {10.1098/rspa.1955.0191},
       adsurl = {https://ui.adsabs.harvard.edu/abs/1955RSPSA.231..505C}
}

@article{hakim_innovative_2023,
	title = {Innovative {Insights} on the {Thin} {Square} {Plate} {Large} {Deflection} {Problem}},
	volume = {16},
	copyright = {http://creativecommons.org/licenses/by/3.0/},
	issn = {1996-1944},
	url = {https://www.mdpi.com/1996-1944/16/21/6967},
	doi = {10.3390/ma16216967},
	language = {en},
	number = {21},
	urldate = {2026-09-14},
	journal = {Materials},
	publisher = {Multidisciplinary Digital Publishing Institute},
	author = {Hakim, Gilad and Abramovich, Haim},
	month = jan,
	year = {2023},
	pages = {6967},
}

@article{ghassemi_comparison_2018,
	title = {Comparison of nonlinear {Von} {Karman} and {Cosserat} theories in very large deformation of skew plates},
	volume = {10},
	issn = {2008-6695},
	url = {https://doi.org/10.1007/s40091-018-0184-2},
	doi = {10.1007/s40091-018-0184-2},
	number = {1},
	journal = {International Journal of Advanced Structural Engineering},
	author = {Ghassemi, Aazam and Hassani, Mohsen and Oveissi, Soheil},
	month = mar,
	year = {2018},
	pages = {73--84},
}

@article{preumont_note_2021,
	title = {A {Note} on the {Electromechanical} {Design} of a {Robotic} {Hummingbird}},
	volume = {10},
	copyright = {http://creativecommons.org/licenses/by/3.0/},
	issn = {2076-0825},
	url = {https://www.mdpi.com/2076-0825/10/3/52},
	doi = {10.3390/act10030052},
	language = {en},
	number = {3},
	urldate = {2026-08-27},
	journal = {Actuators},
	publisher = {Multidisciplinary Digital Publishing Institute},
	author = {Preumont, André and Wang, Han and Kang, Shengzheng and Wang, Kainan and Roshanbin, Ali},
	month = mar,
	year = {2021},
	pages = {52},
}

@article{chung_time_1993,
	title = {A {Time} {Integration} {Algorithm} for {Structural} {Dynamics} {With} {Improved} {Numerical} {Dissipation}: {The} {Generalized}-alpha {Method}},
	volume = {60},
	issn = {0021-8936},
	url = {https://doi.org/10.1115/1.2900803},
	doi = {10.1115/1.2900803},
	number = {2},
	journal = {Journal of Applied Mechanics},
	author = {Chung, J. and Hulbert, G. M.},
	month = jun,
	year = {1993},
	pages = {371--375},
}

@article{crisfield_fast_1981,
	title = {A fast incremental/iterative solution procedure that handles “snap-through”},
	volume = {13},
	issn = {0045-7949},
	url = {https://www.sciencedirect.com/science/article/pii/0045794981901085},
	doi = {https://doi.org/10.1016/0045-7949(81)90108-5},
	number = {1},
	journal = {Computers \& Structures},
	author = {Crisfield, M. A.},
	year = {1981},
	pages = {55--62},
}

@misc{broers_multimodal_2026,
	title = {A {Multimodal} {Tiltwing} {Framework} for {Bioinspired} {Aerial} {Robots}},
	url = {http://arxiv.org/abs/2606.22046},
	doi = {10.48550/arXiv.2606.22046},
	urldate = {2026-08-22},
	publisher = {arXiv},
	author = {Broers, Krispin C. V. and Armanini, Sophie F.},
	month = jun,
	year = {2026},
	note = {arXiv:2606.22046 [cs.RO]},
}

@article{ffowcs_williams_sound_1969,
	title = {Sound generation by turbulence and surfaces in arbitrary motion},
	volume = {264},
	copyright = {https://royalsociety.org/journals/ethics-policies/data-sharing-mining/},
	issn = {0080-4614, 2054-0272},
	url = {https://royalsocietypublishing.org/rsta/article/264/1151/321/44614/Sound-generation-by-turbulence-and-surfaces-in},
	doi = {10.1098/rsta.1969.0031},
	language = {en},
	number = {1151},
	urldate = {2026-08-12},
	journal = {Philosophical Transactions of the Royal Society of London. Series A, Mathematical and Physical Sciences},
	author = {Ffowcs Williams, J. E. and Hawkings, D. L.},
	month = may,
	year = {1969},
	pages = {321--342},
}

@article{rodriguez-perez_dynamic_2000,
	title = {Dynamic mechanical analysis applied to the characterisation of closed cell polyolefin foams},
	volume = {19},
	issn = {01429418},
	url = {https://linkinghub.elsevier.com/retrieve/pii/S0142941899000549},
	doi = {10.1016/S0142-9418(99)00054-9},
	language = {en},
	number = {7},
	urldate = {2026-07-29},
	journal = {Polymer Testing},
	author = {Rodriguez-Perez, M.A and De Saja, J.A},
	month = aug,
	year = {2000},
	pages = {831--848},
}

@article{hashim_insects_2026,
	title = {From insects to bio-inspired {Micro} {Flapping} {Wing} {Aerial} {Vehicles} {Intelligent} {Flight}: {A} review, design principles, and future prospects},
	volume = {11},
	issn = {2950550X},
	shorttitle = {From insects to bio-inspired {Micro} {Flapping} {Wing} {Aerial} {Vehicles} {Intelligent} {Flight}},
	url = {https://linkinghub.elsevier.com/retrieve/pii/S2950550X26000270},
	doi = {10.1016/j.dte.2026.100114},
	language = {en},
	urldate = {2026-07-27},
	journal = {Digital Engineering},
	author = {Hashim, Hashim A.},
	month = dec,
	year = {2026},
	pages = {100114},
}

@article{clark_ways_2021,
	title = {Ways that {Animal} {Wings} {Produce} {Sound}},
	volume = {61},
	copyright = {https://academic.oup.com/journals/pages/open\_access/funder\_policies/chorus/standard\_publication\_model},
	issn = {1540-7063, 1557-7023},
	url = {https://academic.oup.com/icb/article/61/2/696/6162170},
	doi = {10.1093/icb/icab008},
	language = {en},
	number = {2},
	urldate = {2026-07-26},
	journal = {Integrative and Comparative Biology},
	author = {Clark, Christopher J},
	month = sep,
	year = {2021},
	pages = {696--709},
}

@article{reade_investigation_2022,
	title = {Investigation of chordwise functionally graded flexural rigidity in flapping wings using a two-dimensional pitch–plunge model},
	volume = {17},
	issn = {1748-3182, 1748-3190},
	url = {https://iopscience.iop.org/article/10.1088/1748-3190/ac8f05},
	doi = {10.1088/1748-3190/ac8f05},
	language = {en},
	number = {6},
	urldate = {2026-07-26},
	journal = {Bioinspiration \& Biomimetics},
	author = {Reade, Joseph and Jankauski, Mark},
	month = nov,
	year = {2022},
	pages = {066007},
}

@article{hou_comparative_2022,
	title = {Comparative analysis of deformation behaviors of dragonfly wing under aerodynamic and inertial forces},
	volume = {145},
	issn = {00104825},
	url = {https://linkinghub.elsevier.com/retrieve/pii/S001048252200213X},
	doi = {10.1016/j.compbiomed.2022.105421},
	language = {en},
	urldate = {2026-07-26},
	journal = {Computers in Biology and Medicine},
	author = {Hou, Dan and Zhong, Zheng},
	month = jun,
	year = {2022},
	pages = {105421},
}

@article{zhu_parametric_2025,
	title = {Parametric {FEM}-{Based} {Analysis} on {Adaptive} {Vein} {Morphology} of {Dragonfly} {Wings} {Toward} {Material} {Efficiency}},
	volume = {10},
	issn = {2313-7673},
	url = {https://www.mdpi.com/2313-7673/10/12/799},
	doi = {10.3390/biomimetics10120799},
	language = {en},
	number = {12},
	urldate = {2026-07-26},
	journal = {Biomimetics},
	author = {Zhu, Guanqi and Teixeira, Fred Fialho and Loh, Paul},
	month = nov,
	year = {2025},
	pages = {799},
}

@article{combes_flexural_2003,
	title = {Flexural stiffness in insect wings {I}. {Scaling} and the influence of wing venation},
	volume = {206},
	issn = {1477-9145, 0022-0949},
	url = {https://journals.biologists.com/jeb/article/206/17/2979/13853/Flexural-stiffness-in-insect-wings-I-Scaling-and},
	doi = {10.1242/jeb.00523},
	language = {en},
	number = {17},
	urldate = {2026-07-26},
	journal = {Journal of Experimental Biology},
	author = {Combes, S. A. and Daniel, T. L.},
	month = sep,
	year = {2003},
	pages = {2979--2987},
}

@article{combes_into_2003,
	title = {Into thin air: contributions of aerodynamic and inertial-elastic forces to wing bending in the hawkmoth \textit{{Manduca} sexta}},
	volume = {206},
	issn = {1477-9145, 0022-0949},
	shorttitle = {Into thin air},
	url = {https://journals.biologists.com/jeb/article/206/17/2999/13852/Into-thin-air-contributions-of-aerodynamic-and},
	doi = {10.1242/jeb.00502},
	language = {en},
	number = {17},
	urldate = {2026-07-26},
	journal = {Journal of Experimental Biology},
	author = {Combes, S. A. and Daniel, T. L.},
	month = sep,
	year = {2003},
	pages = {2999--3006},
}

@article{wu_design_2026,
	title = {Design and {Experimental} {Validation} of a {Tailless} {Flapping}-{Wing} {Micro} {Aerial} {Vehicle} with {Long} {Endurance} and {High} {Payload} {Capability}},
	volume = {10},
	issn = {2504-446X},
	url = {https://www.mdpi.com/2504-446X/10/1/26},
	doi = {10.3390/drones10010026},
	language = {en},
	number = {1},
	urldate = {2026-06-01},
	journal = {Drones},
	author = {Wu, Chaofeng and Xiao, Yiming and Zhao, Jiaxin and Guo, Qingcheng and Cui, Feng and Wu, Xiaosheng and Liu, Wu},
	month = jan,
	year = {2026},
	pages = {26},
}

@article{avallone_benefits_2017,
	title = {Benefits of curved serrations on broadband trailing-edge noise reduction},
	volume = {400},
	issn = {0022460X},
	url = {https://linkinghub.elsevier.com/retrieve/pii/S0022460X17303140},
	doi = {10.1016/j.jsv.2017.04.007},
	language = {en},
	urldate = {2026-05-30},
	journal = {Journal of Sound and Vibration},
	author = {Avallone, F. and Van Der Velden, W.C.P. and Ragni, D.},
	month = jul,
	year = {2017},
	pages = {167--177},
}

@inproceedings{gruber_experimental_2013,
	address = {Berlin, Germany},
	title = {An experimental investigation of novel trailing edge geometries on airfoil trailing edge noise reduction},
	isbn = {978-1-62410-213-4},
	url = {https://arc.aiaa.org/doi/10.2514/6.2013-2011},
	doi = {10.2514/6.2013-2011},
	language = {en},
	urldate = {2026-05-30},
	booktitle = {19th {AIAA}/{CEAS} {Aeroacoustics} {Conference}},
	publisher = {American Institute of Aeronautics and Astronautics},
	author = {Gruber, Mathieu and Joseph, Phillip and Azarpeyvand, Mahdi},
	month = may,
	year = {2013},
}

@article{truong_effects_2026,
	title = {The effects of bio-inspired wing vein morphology on thrust generation in double-clap flapping-wing robots},
	volume = {55},
	issn = {22149147},
	url = {https://linkinghub.elsevier.com/retrieve/pii/S2214914725002545},
	doi = {10.1016/j.dt.2025.08.003},
	language = {en},
	urldate = {2026-05-17},
	journal = {Defence Technology},
	author = {Truong, Tien Van and Nguyen, Quoc-Viet and Au, Loan Thi Kim and Luong, Hung-Truyen},
	month = jan,
	year = {2026},
	pages = {257--276},
}

@article{debiasi_low-noise_2020,
	title = {Low-{Noise} {Flapping} {Wings} with {Tensed} {Membrane}},
	volume = {58},
	issn = {0001-1452, 1533-385X},
	url = {https://arc.aiaa.org/doi/10.2514/1.J058900},
	doi = {10.2514/1.J058900},
	language = {en},
	number = {6},
	urldate = {2026-04-29},
	journal = {AIAA Journal},
	author = {Debiasi, Marco and Lu, Zhenbo and Nguyen, Quoc Viet and Chan, Woei Leong},
	month = jun,
	year = {2020},
	pages = {2388--2397},
}

@article{ji_influences_2022,
	title = {Influences of serrated trailing edge on the aerodynamic and aeroacoustic performance of a flapping wing during hovering flight},
	volume = {34},
	issn = {1070-6631, 1089-7666},
	url = {https://pubs.aip.org/pof/article/34/1/011902/2845389/Influences-of-serrated-trailing-edge-on-the},
	doi = {10.1063/5.0070450},
	language = {en},
	number = {1},
	urldate = {2026-04-29},
	journal = {Physics of Fluids},
	author = {Ji, Xueyu and Wang, Li and Ravi, Sridhar and Tian, Fang-Bao and Young, John and Lai, Joseph C. S.},
	month = jan,
	year = {2022},
	pages = {011902},
}

@article{wu_experimental_2026,
	title = {Experimental {Investigation} of {Acoustic} {Signatures} in {Hovering} {Flapping}-{Wing} {Micro} {Air} {Vehicle}},
	volume = {64},
	issn = {0001-1452, 1533-385X},
	url = {https://arc.aiaa.org/doi/10.2514/1.J065515},
	doi = {10.2514/1.J065515},
	language = {en},
	number = {3},
	urldate = {2026-04-29},
	journal = {AIAA Journal},
	author = {Wu, Jianghao and Pang, Zepeng and Tian, Zhiming and Zhou, Chao},
	month = mar,
	year = {2026},
	pages = {1367--1380},
}

@inproceedings{keennon_development_2012,
	address = {Nashville, Tennessee},
	title = {Development of the {Nano} {Hummingbird}: {A} {Tailless} {Flapping} {Wing} {Micro} {Air} {Vehicle}},
	isbn = {978-1-60086-936-5},
	shorttitle = {Development of the {Nano} {Hummingbird}},
	url = {https://arc.aiaa.org/doi/10.2514/6.2012-588},
	doi = {10.2514/6.2012-588},
	language = {en},
	urldate = {2026-04-29},
	booktitle = {50th {AIAA} {Aerospace} {Sciences} {Meeting} including the {New} {Horizons} {Forum} and {Aerospace} {Exposition}},
	publisher = {American Institute of Aeronautics and Astronautics},
	author = {Keennon, Matthew and Klingebiel, Karl and Won, Henry},
	month = jan,
	year = {2012},
}

@article{lu_bioinspired_2018,
	title = {Bioinspired {Low}-{Noise} {Wing} {Design} for a {Two}-{Winged} {Flapping}-{Wing} {Micro} {Air} {Vehicle}},
	volume = {56},
	issn = {0001-1452, 1533-385X},
	url = {https://arc.aiaa.org/doi/10.2514/1.J056293},
	doi = {10.2514/1.J056293},
	language = {en},
	number = {12},
	urldate = {2026-04-29},
	journal = {AIAA Journal},
	author = {Lu, Zhenbo and Debiasi, Marco and Nguyen, Quoc-Viet and Chan, Woei-Leong},
	month = dec,
	year = {2018},
	pages = {4697--4705},
}

@article{stanford_aeroelastic_2012,
	title = {Aeroelastic {Optimization} of {Flapping} {Wing} {Venation}: {A} {Cellular} {Division} {Approach}},
	volume = {50},
	issn = {0001-1452, 1533-385X},
	shorttitle = {Aeroelastic {Optimization} of {Flapping} {Wing} {Venation}},
	url = {https://arc.aiaa.org/doi/10.2514/1.J051443},
	doi = {10.2514/1.J051443},
	language = {en},
	number = {4},
	urldate = {2026-04-29},
	journal = {AIAA Journal},
	author = {Stanford, Bret and Beran, Philip and Kobayashi, Marcelo},
	month = apr,
	year = {2012},
	pages = {938--951},
}

@article{rafee_nekoo_review_2025,
	title = {A review on flapping-wing robots: {Recent} progress and challenges},
	volume = {44},
	issn = {0278-3649, 1741-3176},
	shorttitle = {A review on flapping-wing robots},
	url = {https://journals.sagepub.com/doi/10.1177/02783649251343638},
	doi = {10.1177/02783649251343638},
	language = {en},
	number = {14},
	urldate = {2026-04-29},
	journal = {The International Journal of Robotics Research},
	author = {Rafee Nekoo, Saeed and Rashad, Ramy and De Wagter, Christophe and Fuller, Sawyer B and Croon, Guido De and Stramigioli, Stefano and Ollero, Anibal},
	month = dec,
	year = {2025},
	pages = {2305--2339},
}

\end{document}